\documentclass[10pt,a4paper]{article}
\usepackage[T1]{fontenc}
\usepackage[utf8]{inputenc}
\usepackage{lmodern}
\usepackage[margin=0.82in,top=0.78in,bottom=0.80in,headheight=14pt]{geometry}
\usepackage{amsmath,amssymb,booktabs,array,tabularx,microtype}
\usepackage[table]{xcolor}
\usepackage{graphicx,tikz,placeins}
\graphicspath{{assets/}}
\usetikzlibrary{arrows.meta,positioning,fit,backgrounds}
\usepackage{enumitem,caption,fancyhdr,titlesec,hyperref,needspace}
\definecolor{navy}{HTML}{24364B}
\definecolor{teal}{HTML}{426887}
\definecolor{pale}{HTML}{EDF6F7}
\definecolor{graytext}{HTML}{586570}
\hypersetup{colorlinks=true,linkcolor=teal,citecolor=teal,urlcolor=teal,
  pdftitle={Axiom-0: Hierarchical Self-Refinement for Vision-Language-Action Policies — Toward Safe Recursive Self-Improvement in Embodied Systems},
  pdfauthor={4Axiom Robotics Team},
  pdfsubject={Axiom-0 Technical Report}}
\titleformat{\section}{\large\bfseries\color{navy}}{\thesection}{0.7em}{}
\titleformat{\subsection}{\normalsize\bfseries\color{navy}}{\thesubsection}{0.7em}{}
\titlespacing*{\section}{0pt}{12pt}{7pt}
\titlespacing*{\subsection}{0pt}{10pt}{5pt}
\setlist{nosep,leftmargin=1.4em,topsep=4pt}
\newcommand{\model}{Axiom-0}
\newcommand{\data}{\texttt{pretrain\_human300}}

\begin{document}
\thispagestyle{plain}
\noindent\begin{minipage}[c]{0.30\linewidth}
\includegraphics[width=3.6cm]{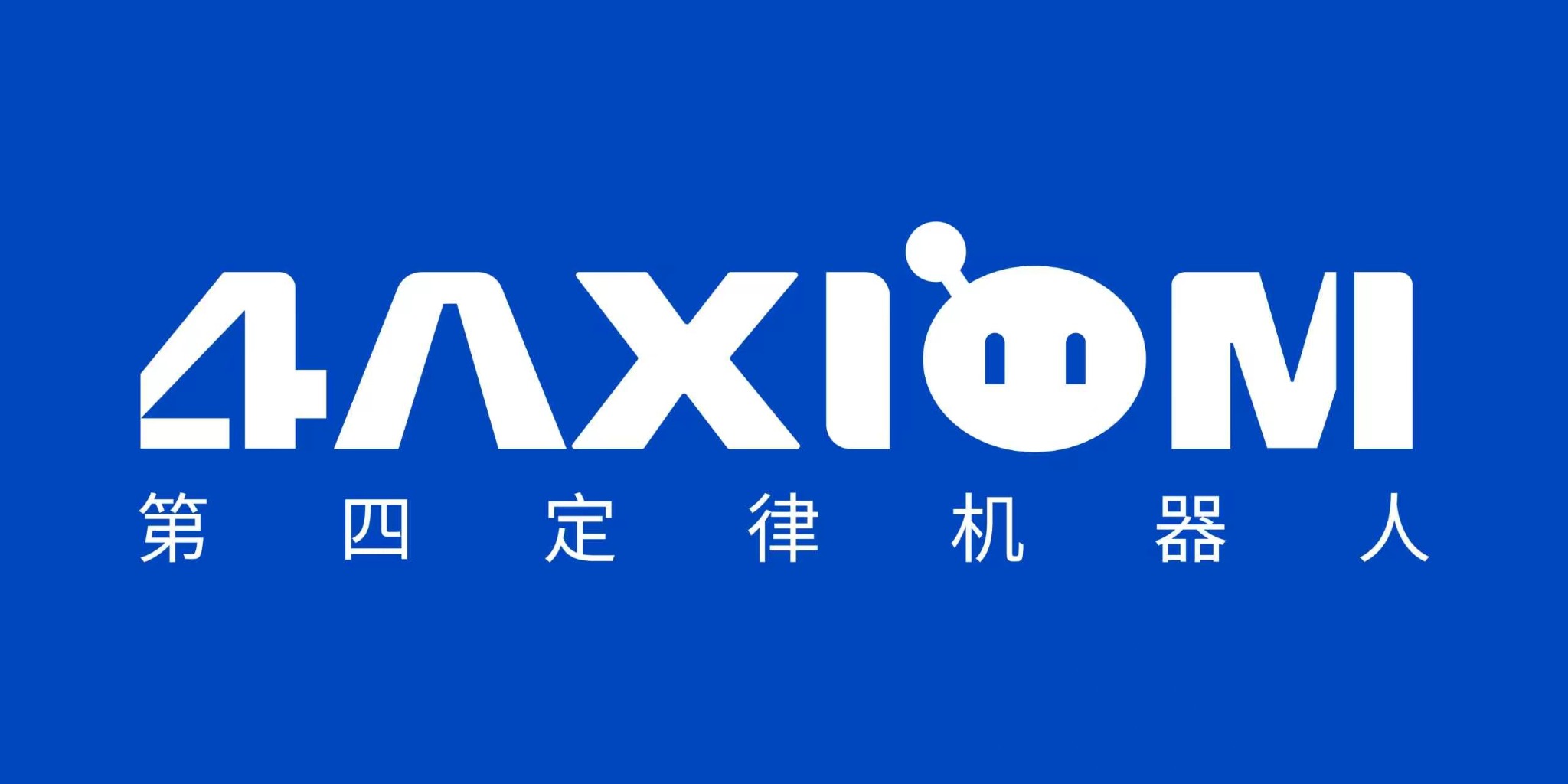}
\end{minipage}\hfill
\begin{minipage}[c]{0.65\linewidth}
\raggedleft\small\bfseries\color{teal}TECHNICAL REPORT
\end{minipage}
\vspace{5pt}
\begin{center}
{\LARGE\bfseries\color{navy}Axiom-0: Hierarchical Self-Refinement\\[4pt]
for Vision-Language-Action Policies}\par
\vspace{6pt}
{\normalsize\itshape\color{graytext}Our First Step Toward Safe Recursive Self-Improvement in Embodied Systems}\par
\vspace{8pt}
Renbin Shu, Jing Yang, Xiangyuan You, Guolin Qiu, Yufei Xia\\
Yiming Zhu, Nanjun Dai, Fei Zhang, Feng Wu, Keke Tang\\

{\large 4Axiom Robotics Team}\par
\vspace{4pt}
{\small October 9, 2026}
\end{center}
\vspace{4pt}
\begin{abstract}
\noindent
Reliable manipulation requires policies that generalize across tasks, layouts, and object interactions. Vision-language-action (VLA) models combine pretrained multimodal understanding with learned control, but their performance remains limited on unfamiliar task compositions. We present \model, a VLA policy that addresses both the quality of its representations and their use in action generation through hierarchical self-refinement. Intra-VLM Representation Refinement (IVRR) strengthens selected intermediate features through consistency supervision with learned feature projections during training. VLM--Action Interface Refinement (VAIR) recursively updates the resulting conditioning before action generation. Joint optimization of the supervised VLM layers, VAIR, and the action generator connects the two components. On RoboCasa365, \model{} achieves 62.32\% overall success in our evaluation, exceeding the highest published leaderboard score (Paimon-0, 58.1\%) by 4.22 percentage points and the Xiaomi-Robotics-1 reference by 4.92 points. Under the evaluated profiling setup, mean action-chunk generation latency increases by 1.7\% over Xiaomi-Robotics-1. These results demonstrate improved task completion with modest inference overhead.
\end{abstract}

\section{Introduction}
Robots operating in everyday environments must complete tasks across changing objects, layouts, and instructions. This requires more than recognizing a scene: a policy must retain the information needed for physical interaction and use it to generate appropriate actions. Vision-language-action (VLA) models bring multimodal understanding and control into a shared learning framework. Improving the reliability of this connection is central to deploying robots beyond familiar training conditions.

Recent systems such as $\pi_0$, $\pi_{0.5}$, and Xiaomi-Robotics-1 combine pretrained vision-language models (VLMs) with generative action models to broaden their manipulation capabilities~\cite{pi0,pi05,xiaomi}. Evaluation across a wider range of conditions also exposes their limitations. RoboCasa365 tests generalization to unseen task compositions~\cite{robocasa}, LIBERO-Plus examines sensitivity to controlled perturbations~\cite{liberoplus}, and LIBERO-Safety evaluates physical and semantic safety failures~\cite{liberosafety}. These settings place demands on both scene understanding and control: relevant information must be represented, and the action generator must use it effectively. Strong performance in familiar settings does not resolve either requirement under distribution shift.

We study these requirements as two coupled parts of policy learning. The first concerns the information retained in intermediate multimodal representations. The second concerns how those representations condition action generation. Supervising the first does not, by itself, specify how the second should exploit the additional information. Our design therefore couples representation supervision with learned refinement at the VLM--action interface. The supervised layers continue to adapt with the interface and action generator, allowing the representation and its downstream use to be optimized together. This yields a hierarchical \emph{refine-before-act} policy.

In this report, we introduce \model, the first model in 4Axiom Robotics' research program toward Safe Recursive Self-Improvement (SRSI) in embodied systems. \emph{Intra-VLM Representation Refinement (IVRR)} supervises selected intermediate features through learned projections, using spatial foundation-model references in this version. \emph{VLM--Action Interface Refinement (VAIR)} applies a shared recurrent block to the conditioning values consumed by a diffusion Transformer (DiT). Selective joint optimization connects the two components. We evaluate the resulting policy on RoboCasa365 through comparisons with published methods, successive model configurations, and inference profiling. The longer-term role of feedback-driven policy improvement is discussed in Section~\ref{sec:limitations}.

\section{Method}\label{sec:method}
\subsection{Problem Formulation and Overview}\label{sec:overview}
At decision time $t$, the policy receives multi-view observations $o_t$, an instruction $\ell$, and robot state $s_t$, and predicts an action chunk $A_t=(a_t,\ldots,a_{t+H-1})$ of horizon $H$, represented in model action coordinates as $A_t\in\mathbb R^{H\times d_a}$, where $d_a$ is the action dimension. We consider a VLM--DiT policy in which a VLM $F_\theta$ produces multimodal conditioning $\mathcal C_t$ and a DiT $D_\psi$ predicts the velocity field used for flow-matching action generation:
\begin{equation}
 \mathcal C_t=F_\theta(o_t,\ell),\qquad
 v_\psi(x_\tau,\tau\mid\mathcal C_t,s_t)
 =D_\psi(x_\tau,\tau;\mathcal C_t,s_t).
 \label{eq:backbone}
\end{equation}
Here $\mathcal C_t$ contains image--language conditioning, while robot state $s_t$ enters the DiT separately. The flow state $x_\tau\in\mathbb R^{H\times d_a}$ follows the noise-to-action convention of the underlying action model~\cite{xiaomi,xrrepo}. For a training action chunk $A_t^{\mathrm{gt}}$ and noise $\epsilon\sim\mathcal N(0,I)$ of the same shape, the interpolation and target velocity are
\begin{equation}
 x_\tau=(1-\tau)\epsilon+\tau A_t^{\mathrm{gt}},\qquad
 u_\tau=\frac{\partial x_\tau}{\partial\tau}=A_t^{\mathrm{gt}}-\epsilon,
 \quad \tau\in[0,1].
 \label{eq:flow-path}
\end{equation}
The flow-matching term trains $v_\psi$ to predict this velocity on the interpolated actions. At inference, the five-step Euler sampler draws fresh Gaussian noise, holds the conditioning fixed, and computes
\begin{equation}
 \begin{aligned}
 x_0&=\epsilon,\qquad
 x_{(k+1)/K}=x_{k/K}+\frac{1}{K}
 v_\psi(x_{k/K},k/K\mid\mathcal C_t,s_t),\\
 A_t&=x_1=\operatorname{Sample}_{D_\psi}(\mathcal C_t,s_t;\epsilon),
 \qquad k=0,\ldots,K-1,\quad K=5.
 \end{aligned}
 \label{eq:base-sampling}
\end{equation}
The Gaussian is over the vectorized action chunk; decoding to robot commands follows sampling.
This formulation exposes two locations for refinement: the intermediate representations learned by $F_\theta$ and the conditioning $\mathcal C_t$ supplied to $D_\psi$.

\begin{figure}[!htbp]
\centering
\includegraphics[width=.98\linewidth]{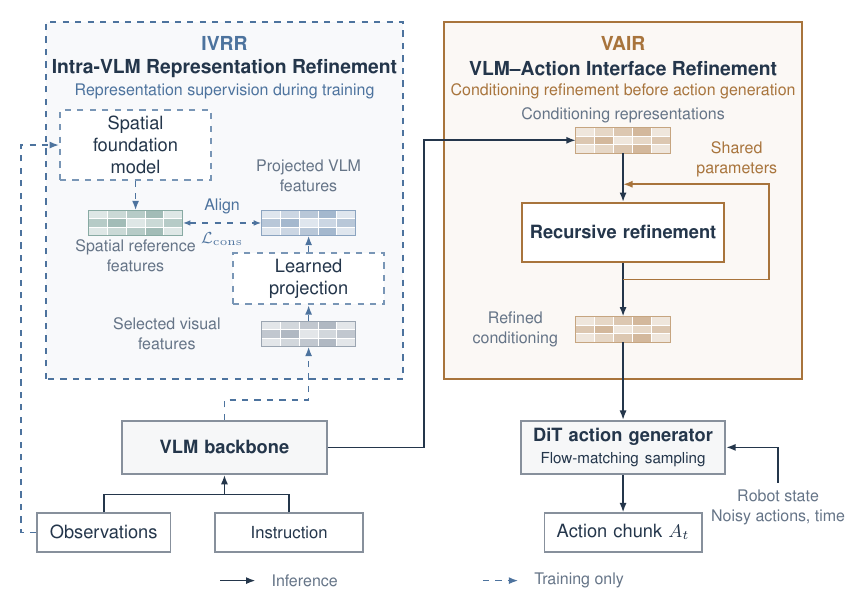}
\caption{\textbf{Overview of hierarchical self-refinement in Axiom-0.} \textbf{Left: IVRR} aligns projected intermediate VLM features with spatial foundation-model references during training. \textbf{Right: VAIR} recursively refines the conditioning supplied to the action generator using a shared refinement module. Common inputs, the VLM backbone, and the DiT action generator lie outside the refinement panels. Solid arrows denote inference paths; dashed arrows denote auxiliary training paths. Feature grids are schematic. Figure~\ref{fig:vair} expands the VAIR computation and its integration with action sampling.}
\label{fig:architecture}
\end{figure}

Figure~\ref{fig:architecture} summarizes the hierarchy: IVRR supervises selected intermediate VLM features during training, and VAIR refines their conditioning output before action sampling. Axiom-0 instantiates this hierarchy on the VLM--DiT architecture of Xiaomi-Robotics-1~\cite{xiaomi}; implementation and checkpoint provenance are specified in Section~\ref{sec:setup}. We first describe each component, then explain their composition, training, and inference in Section~\ref{sec:composition}.

\Needspace{12\baselineskip}
\subsection{Intra-VLM Representation Refinement}\label{sec:ivrr}
IVRR supervises intermediate VLM features alongside action learning. In this version, a pretrained spatial foundation model provides reference features from the training observations, and learned projections connect selected VLM features to these references. IVRR draws on spatially informed representation learning~\cite{spatialvla,falcon} and foundation-model feature alignment~\cite{spatialforcing,vega}. These supervised representations subsequently supply the conditioning refined by VAIR.

\Needspace{12\baselineskip}
\paragraph{Spatial reference and learned projection.}
Let $E_{\mathrm{sp}}$ denote the spatial foundation model, $H_{t,\mathrm{vis}}^l$ the visual features at VLM layer $l$, and $\mathcal J$ the nonempty supervised layer set. For the same observation $o_t$, the auxiliary training branch forms
\begin{equation}
 T_t=E_{\mathrm{sp}}(o_t),\qquad
 \widehat H_t^l=G_l(H_{t,\mathrm{vis}}^l),\quad l\in\mathcal J,
 \label{eq:spatial-reference}
\end{equation}
where $G_l$ is a learnable neural projection that maps the selected VLM features into a space compatible with the spatial reference. Alignment compares projected VLM features with spatial references from the same observations; it does not equate the models' raw hidden states. The spatial model supplies a training signal; its features are not additional inputs to the deployed action generator.

\paragraph{Consistency supervision.}
Writing $\mathcal D_l$ for the feature-alignment discrepancy at layer $l$, including spatial correspondence and reduction over the aligned visual positions, the representation objective and the first-stage training objective are
\begin{align}
 \mathcal L_{\mathrm{cons}} & =
 \frac{1}{|\mathcal J|}\sum_{l\in\mathcal J}
 \mathcal D_l\!\left(\widehat H_t^l,T_t\right),
 \label{eq:alignment}\\
 \mathcal L_{\mathrm{I}} & =
 \mathcal L_{\mathrm{action}}+\lambda\,\mathcal L_{\mathrm{cons}},
 \label{eq:intra}
\end{align}
with losses averaged over training samples. The action objective $\mathcal L_{\mathrm{action}}$ includes DiT velocity regression using Equation~\ref{eq:flow-path}; $\lambda\geq0$ weights the added supervision. Jointly training the projections and VLM connects spatial alignment to the action objective.

After alignment training, the supervised VLM layers continue to adapt jointly with VAIR and the DiT (Section~\ref{sec:composition}). The spatial foundation model and alignment projections are used only during training; deployment uses the learned VLM features without the auxiliary alignment branch.

\subsection{VLM--Action Interface Refinement}\label{sec:vair}
VAIR refines the VLM conditioning before action sampling. At each adapted interface $l\in\mathcal I$, it receives the conditioning keys and values $(K_C^l,V_C^l)$ and predicts a residual correction to the values. Keys support context addressing, while values supply the content aggregated by action queries. Refining values provides a direct way to alter that content while retaining the incoming keys and the structure of the pretrained interface.

\begin{figure}[!htbp]
\centering
\includegraphics[width=.94\linewidth]{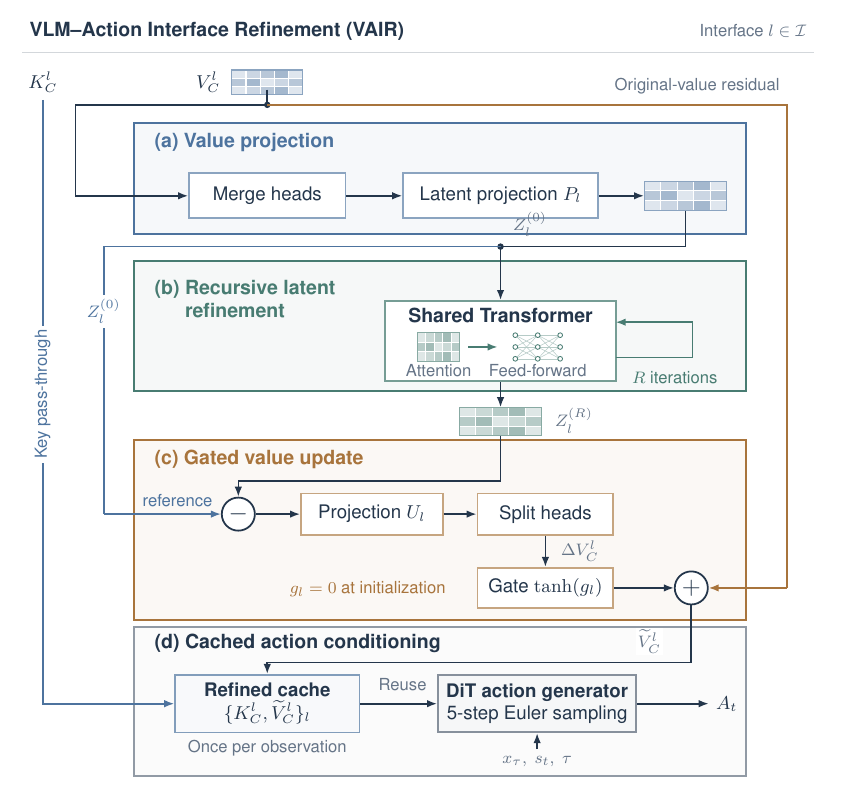}
\caption{\textbf{VAIR computation and integration with the action generator.} \textbf{(a)--(c)} show the added refinement computation: values are projected into a compact latent space, updated by a shared Transformer over $R$ iterations, and converted into a gated residual correction to the original values. The subtraction in (c) takes $Z_l^{(R)}$ as its input and $Z_l^{(0)}$ as its reference. Keys pass through unchanged. \textbf{(d)} shows the refined cache and the DiT sampling path; the cache is computed once per observation and reused across five Euler sampling steps. Feature grids and Transformer internals are schematic.}
\label{fig:vair}
\end{figure}

Figure~\ref{fig:vair} expands the high-level refinement loop in Figure~\ref{fig:architecture} into four operations: \emph{value projection}, \emph{recursive latent refinement}, \emph{gated value update}, and \emph{cached action conditioning}. The first three implement the added VAIR computation; the fourth shows its integration with the DiT action generator. The adapted interface set $\mathcal I$ identifies where VAIR is applied, whereas $\mathcal J$ identifies the VLM layers supervised and subsequently jointly optimized.

\Needspace{9\baselineskip}
\paragraph{(a) Value projection.}
We merge the incoming value heads and apply a learned projection $P_l$ into a compact latent space:
\begin{equation}
 Z_l^{(0)}=P_l\!\left(\operatorname{MergeHeads}(V_C^l)\right),
 \qquad Z_l^{(0)}\in\mathbb R^{B\times N\times d_b}.
 \label{eq:projection}
\end{equation}
Here $B$ is the batch size, $N$ the number of conditioning tokens, and $d_b$ the latent width. For $h_{\mathrm{kv}}$ value heads of width $d_h$, $V_C^l\in\mathbb R^{B\times h_{\mathrm{kv}}\times N\times d_h}$. Head merging produces width $d_v=h_{\mathrm{kv}}d_h$, and $P_l:\mathbb R^{d_v}\to\mathbb R^{d_b}$ acts token-wise. Projection reduces the channel width while retaining the conditioning-token sequence for recurrent processing.

\Needspace{10\baselineskip}
\paragraph{(b) Recursive latent refinement.}
A looped Transformer~\cite{looped} repeatedly updates the projected state:
\begin{equation}
 Z_l^{(r+1)}=T_{\phi_l}(Z_l^{(r)}),\qquad r=0,\ldots,R-1.
 \label{eq:recurrence}
\end{equation}
The block contains attention, feed-forward layers, and residual connections. Parameters are shared across the $R$ iterations at a given interface. The refined state $Z_l^{(R)}$ results from repeated token interaction, with no additional block parameters as $R$ increases. Its difference from the initial state determines the value correction.

\paragraph{(c) Gated value update.}
The difference between the refined and initial latent states defines a correction to the original values:
\begin{align}
 \Delta V_C^l &= \operatorname{SplitHeads}\!\left(U_l(Z_l^{(R)}-Z_l^{(0)})\right),\label{eq:delta}\\
 \widetilde V_C^l &= V_C^l+\tanh(g_l)\,\Delta V_C^l,
 \qquad \widetilde K_C^l=K_C^l.\label{eq:residual}
\end{align}
The token-wise projection $U_l:\mathbb R^{d_b}\to\mathbb R^{d_v}$ restores the value-channel dimension, and $\operatorname{SplitHeads}$ inverts the head-merging reshape. Thus $\Delta V_C^l$ has the same shape as $V_C^l$; the gate $\tanh(g_l)$ is broadcast over that tensor. Initializing $g_l=0$ makes the adapter initially preserve the incoming cache. At interfaces outside $\mathcal I$, keys and values pass through unchanged. Key pass-through is a property of VAIR's forward operation: the selected VLM layers remain trainable, and their output keys and values can both change during optimization.

\Needspace{12\baselineskip}
\paragraph{(d) Cached action conditioning.}
The refined cache supplies context to the DiT. For one attention head, let $Q_A^l\in\mathbb R^{H\times d_h}$ denote the action-query rows and $K_D^l,V_D^l\in\mathbb R^{L_D\times d_h}$ the keys and values of the full DiT token stream, including state, noisy-action, and auxiliary sink tokens. With batch indices suppressed, the action-row output before head concatenation and output projection is
\begin{equation}
 O_A^l=\operatorname{softmax}_{\mathrm{keys}}\!\left(
 \frac{Q_A^l[K_C^l;K_D^l]^\top}{\sqrt{d_h}}+M_A^l
 \right)[\widetilde V_C^l;V_D^l].
 \label{eq:attention}
\end{equation}
Here each conditioning head $K_C^l,\widetilde V_C^l\in\mathbb R^{N\times d_h}$ is selected or repeated to match the query head, and $[\,;\,]$ concatenates tokens. Queries and keys include the backbone's normalization and positional transformations. The additive mask $M_A^l\in\mathbb R^{H\times(N+L_D)}$ is zero at allowed positions and $-\infty$ at blocked positions; softmax is row-wise over keys. The output therefore has shape $H\times d_h$. VAIR runs once per observation update. The same refined cache is reused throughout the DiT's five-step Euler sampler, so latent refinement is not repeated at every sampling step. The additional recurrent computation is therefore paid once per action chunk.

\subsection{Training and Inference}\label{sec:composition}
\paragraph{Policy integration.}
IVRR and VAIR address successive parts of the policy defined in Section~\ref{sec:overview}. IVRR shapes the representations learned by $F_\theta$, whose conditioning output is instantiated as the layer-wise cache $\mathcal C_t=\{K_C^l,V_C^l\}_l$. VAIR, denoted by $\mathcal S_\phi$ with parameters $\phi$ collecting its projections, recurrent blocks, and gates, refines this cache before it enters the action sampler. The complete policy replaces the conditioning in Equation~\ref{eq:base-sampling} with
\begin{equation}
 \widetilde{\mathcal C}_t=\mathcal S_\phi(\mathcal C_t),\qquad
 A_t=\operatorname{Sample}_{D_\psi}(\widetilde{\mathcal C}_t,s_t;\epsilon).
 \label{eq:policy}
\end{equation}
Here IVRR is realized in the trained parameters $\theta$, while $\mathcal S_\phi$ remains an explicit computation in the inference path.

\paragraph{Training.}
Training proceeds in two stages. Stage I optimizes the full VLM and DiT together with the alignment projections using Equation~\ref{eq:intra}, applying consistency supervision at layers $\mathcal J$. Stage II jointly optimizes those VLM layers, VAIR, and the DiT under the action objective, while holding the remaining VLM layers fixed. This second stage adapts the supervised representations to their use by the refined action interface. Training configuration and data scope are given in Section~\ref{sec:setup}.

\paragraph{Inference.}
All model parameters are fixed during deployment. For each new observation, the VLM produces its conditioning cache and VAIR computes the refined values once. The DiT reuses this cache throughout action sampling to produce an action chunk. Neither auxiliary alignment nor parameter updates are required at inference.

\section{Experiments}\label{sec:experiments}
We evaluate task completion, compare the successive model configurations, and measure the inference cost of the complete policy.

\subsection{Experimental Setup}\label{sec:setup}
\paragraph{Benchmark and metric.}
We use the RoboCasa365 target-50 multi-task evaluation suite with RoboCasa 1.0.1~\cite{robocasa,leaderboard}. It contains 18 Atomic-Seen, 16 Composite-Seen, and 16 Composite-Unseen tasks. We evaluate 50 episodes per task, giving 2,500 episodes in total. Task success is reported as a percentage for each split and over all 50 tasks. With equal episode counts per task, overall success equals the fraction of successful episodes. Score differences are expressed in percentage points (pp).

\paragraph{Implementation and initialization.}
Axiom-0 retains the VLM backbone and DiT action-generator architecture of Xiaomi-Robotics-1 and initializes these components from its publicly released checkpoint~\cite{xiaomi,xrrepo}. We augment this architecture with IVRR training supervision and the VAIR conditioning-refinement module, and optimize the resulting policy using the two-stage procedure in Section~\ref{sec:composition}.

\paragraph{Training data.}
All subsequent RoboCasa-specific training uses only \data{}. No target-50 demonstrations, evaluation trajectories, or additional training datasets are used. Auxiliary spatial reference features are computed from these same training observations using the pretrained spatial foundation model. This dataset restriction concerns RoboCasa-specific training; pretrained components retain their upstream training provenance.

\Needspace{6\baselineskip}
\paragraph{Training configuration.}
Stage I runs for 16,000 optimization steps, updating all VLM and DiT parameters with alignment supervision at the selected VLM layers. Stage II continues for 2,000 steps, updating those layers together with VAIR and the DiT. The global batch size is 2,048. Training uses 32 NVIDIA RTX PRO 5000 GPUs with 72\,GB memory each, distributed across four nodes.

\paragraph{Evaluation and comparison scope.}
Our success rates are evaluated by the authors and are pending independent benchmark verification. Published comparator scores follow the official leaderboard snapshot accessed October 3, 2026~\cite{leaderboard}. They provide benchmark references across reported systems and training configurations. The component comparison uses the published Xiaomi-Robotics-1 result as its base reference. Repeated-run uncertainty estimates and matched-budget component controls are not available for the present report.

\subsection{Comparison with Published Methods}
Table~\ref{tab:sota} reports the comparison with published RoboCasa365 methods. \model{} succeeds in 1,558 of 2,500 episodes, giving 62.32\% overall success. This is 4.92 points above Xiaomi-Robotics-1 and 4.22 points above Paimon-0, the strongest published overall result in the snapshot. Split scores retain the precision of our submission, and the overall result is computed from episode outcomes.

\begin{table}[!htbp]
\centering\small
\caption{\textbf{RoboCasa365 benchmark comparison.} Success rates (\%), using the published leaderboard snapshot of October 3, 2026~\cite{leaderboard}. Ours is author-evaluated. Bold marks the largest reported point estimate in each column.}
\label{tab:sota}
\setlength{\tabcolsep}{6pt}
\renewcommand{\arraystretch}{1.15}
\begin{tabularx}{\linewidth}{@{}Xrrrr@{}}
\toprule
Method & \shortstack{Atomic\\Seen} & \shortstack{Composite\\Seen} & \shortstack{Composite\\Unseen} & Overall\\
\midrule
Paimon-0 & 79.3 & 55.8 & 36.7 & 58.1\\
Xiaomi-Robotics-1 & 80.2 & 57.1 & 32.1 & 57.4\\
Phasor-m7 & 78.7 & 57.4 & 31.5 & 56.8\\
ProWAM & 76.0 & 46.5 & 18.2 & 48.1\\
PatchWAM & 78.7 & 46.4 & 13.8 & 47.6\\
ABot-M0.6 & 79.4 & 48.3 & 7.9 & 46.6\\
ABot-M0.5 & 75.6 & 37.7 & 3.3 & 40.3\\
PRTS & 66.3 & 30.3 & 18.8 & 39.6\\
RLDX-1 & 67.6 & 27.9 & 8.5 & 36.0\\
WorldDreamer & 66.3 & 26.7 & 9.0 & 35.3\\
GR00T N1.5 & 50.7 & 14.8 & 2.7 & 23.9\\
LY-GWM-RoboCasa & 48.3 & 16.3 & 3.1 & 23.6\\
GR00T N1.6 & 51.1 & 9.4 & 1.7 & 21.9\\
GigaWorld-Policy 0.1 & 44.4 & 11.8 & 2.9 & 20.7\\
$\pi_{0.5}$ & 39.6 & 7.1 & 1.2 & 16.9\\
$\pi_0$ & 34.6 & 6.1 & 1.1 & 14.8\\
Diffusion Policy & 15.7 & 0.2 & 1.3 & 6.1\\
\midrule
Ours (Axiom-0) & \textbf{86.22} & \textbf{60.12} & \textbf{37.62} & \textbf{62.32}\\
\bottomrule
\end{tabularx}
\end{table}

The improvement over Xiaomi-Robotics-1 is observed on all three splits: +6.02 points on Atomic-Seen, +3.02 on Composite-Seen, and +5.52 on Composite-Unseen. Relative to Paimon-0~\cite{paimon}, the corresponding differences are +6.92, +4.32, and +0.92 points. Thus, the comparison supports an overall task-completion gain, with a smaller margin over the strongest comparator on unseen compositions. These point estimates do not establish statistical significance.

\FloatBarrier
\subsection{Component Analysis}

Table~\ref{tab:ablation} compares four configurations: the published
Xiaomi-Robotics-1 reference, IVRR only, VAIR only, and the complete
Axiom-0 policy. The individual configurations assess whether either
component can improve task completion without the other, while the
complete model evaluates their combined use.

\begin{table}[!htbp]
\centering
\small
\caption{\textbf{Component configurations on RoboCasa365.}
IVRR: Intra-VLM Representation Refinement;
VAIR: VLM--Action Interface Refinement.
Success rates are reported in \%; $\Delta$ denotes the overall gain
in percentage points over the published Xiaomi-Robotics-1
reference~\cite{xrsubmission}.
Split scores retain their reported precision.
Comparisons reflect the resulting trained configurations.}
\label{tab:ablation}
\setlength{\tabcolsep}{5pt}
\renewcommand{\arraystretch}{1.15}
\begin{tabularx}{\linewidth}{@{}cc*{5}{>{\raggedleft\arraybackslash}X}@{}}
\toprule
IVRR & VAIR
& \shortstack{Atomic\\Seen}
& \shortstack{Composite\\Seen}
& \shortstack{Composite\\Unseen}
& Overall & $\Delta$ \\
\midrule
-- & --
& 80.20 & 57.10 & 32.10 & 57.40 & -- \\
$\checkmark$ & --
& 84.11 & 56.38 & 37.50 & 60.32 & +2.92 \\
-- & $\checkmark$
& 83.78 & 60.125 & 35.50 & 60.76 & +3.36 \\
$\checkmark$ & $\checkmark$
& \textbf{86.22} & 60.12 & \textbf{37.62}
& \textbf{62.32} & \textbf{+4.92} \\
\bottomrule
\end{tabularx}
\end{table}

IVRR alone raises overall success from 57.40\% to 60.32\%, a gain
of 2.92 percentage points. Its largest improvement is on
Composite-Unseen, where success increases from 32.10\% to 37.50\%.
Atomic-Seen also improves, whereas Composite-Seen decreases by
0.72 points. This configuration therefore shows its strongest
benefit on generalization to unseen task compositions.

VAIR alone achieves 60.76\% overall success, exceeding the base
reference by 3.36 points, with improvements across all three
splits. These results show that the interface-refinement
configuration improves task completion even without IVRR
supervision. Relative to IVRR alone, VAIR alone performs better
on Composite-Seen but less well on Composite-Unseen, indicating
different performance profiles for the two components.

The complete model achieves the highest overall success of
62.32\%, improving over IVRR alone by 2.00 points and VAIR alone
by 1.56 points. Compared with VAIR alone, it gains 2.44 points
on Atomic-Seen and 2.12 points on Composite-Unseen, while
Composite-Seen performance remains approximately 60.1\%.
Thus, the complete configuration retains VAIR's Composite-Seen
performance while improving the other two splits.

The results are consistent with complementary roles for
representation supervision and conditioning refinement.
They establish that both individual configurations outperform
the published reference and that their combined use yields
the strongest overall result. Because these comparisons also
involve optimization of the backbone and action generator,
matched-budget controls and repeated evaluations remain
necessary to isolate architectural effects and quantify
variability.

\subsection{Computational Cost and Inference Efficiency}\label{sec:efficiency}
We profile the complete \model{} policy and Xiaomi-Robotics-1 on an NVIDIA RTX PRO 5000 (72\,GB), with batch size one and 16 actions per generated chunk. Table~\ref{tab:efficiency} reports generation latency, model parameters, estimated dominant floating-point operations, and peak GPU memory. Latency measures computation for one complete action chunk, excluding its physical execution. FLOPs denote work per chunk rather than hardware throughput.

\begin{table}[!htbp]
\centering\small
\caption{\textbf{Inference cost on an NVIDIA RTX PRO 5000 (72\,GB).} Batch size is one; each call produces a 16-action chunk. B and M denote billions and millions of parameters.}
\label{tab:efficiency}
\setlength{\tabcolsep}{4pt}
\renewcommand{\arraystretch}{1.20}
\begin{tabularx}{\linewidth}{@{}Xrrr@{}}
\toprule
Metric & Xiaomi-Robotics-1 & Ours & Change\\
\midrule
Mean latency (ms) & 133.25 & 135.55 & +2.30 (1.73\%)\\
Median latency (ms) & 131.76 & 134.78 & +3.02 (2.29\%)\\
P95 latency (ms) & 139.19 & 140.53 & +1.34 (0.96\%)\\
Parameters (B) & 5.11322 & 5.12791 & +14.684\,M (0.287\%)\\
Estimated compute (TFLOPs/chunk) & 4.751 & 4.779 & +0.028 (0.59\%)\\
Peak GPU memory (GiB) & 9.709 & 9.763 & +0.054 (0.56\%)\\
\bottomrule
\end{tabularx}
\vspace{4pt}
\begin{minipage}{\linewidth}\footnotesize
Each change is an absolute increment followed by its percentage relative to Xiaomi-Robotics-1. Increments use the metric units unless marked otherwise. Rounded totals and reported increments may differ in their final digits. Compute estimates cover dominant operations and denote work per chunk, not hardware throughput.
\end{minipage}
\end{table}

Mean generation latency increases from 133.25 to 135.55\,ms, an additional 2.30\,ms per chunk (1.73\%). P95 latency increases by 1.34\,ms. The added parameters total approximately 14.7 million (0.287\% of the base model), estimated computation rises by approximately 0.6\%, and peak GPU memory increases by approximately 55\,MiB. Together with the 4.92-point overall success improvement, these measurements indicate a favorable success--cost trade-off for the evaluated configuration.


\FloatBarrier
\Needspace{14\baselineskip}
\section{Discussion and Future Directions}\label{sec:limitations}
\paragraph{What the present results establish.}
Axiom-0 connects representation supervision with recurrent
conditioning refinement through selective joint optimization.
The component study now includes both IVRR-only and VAIR-only
configurations. Each improves overall success over the published
Xiaomi-Robotics-1 reference, and the complete policy performs
best overall. In particular, the VAIR-only result shows that
the improvement associated with interface refinement does not
require IVRR supervision. Further controlled experiments will
examine how the gains depend on recurrence depth, the choice
of supervised VLM layers, and joint optimization. Comparisons
with non-recurrent adapters under matched parameter and
training budgets, together with repeated evaluations, will
help distinguish the benefit of recurrent computation from
additional capacity and optimization. Profiling remains
limited to one device and workload; warm-up counts, numerical
precision, and detailed timing boundaries are not recorded
in the available measurements.

\paragraph{From self-refinement to SRSI.}
Our long-term objective is a robot that improves from experience while preserving dependable behavior. Improvement would be recursive: each updated policy would generate new experience, reveal further limitations, and inform subsequent updates. Axiom-0 addresses an earlier and narrower problem. It refines conditioning with fixed parameters before acting; it does not yet diagnose failures, revise its own policy, or validate and adopt updates autonomously. Task-completion results support the present design, but do not establish the safety or effectiveness of such a learning cycle.

\paragraph{Learning from execution feedback.}
A next step is to connect refinement to evidence from action outcomes. We plan to study how discrepancies between intended and observed outcomes can identify missing information in the representation, errors in its use by the action generator, and limitations of the task strategy itself. Candidate supervision signals include temporal dynamics, task progress, and scene uncertainty. We will investigate whether this feedback can guide subsequent representation and policy updates, as well as determine when additional inference-time refinement is useful. 

\paragraph{Reliable improvement over successive updates.}
For SRSI, updates must remain useful beyond the experience that motivated them. We will study how to evaluate candidate updates across held-out tasks, distribution shifts, recovery scenarios, and existing skills before adoption. We also plan to study how to retain prior capabilities, detect regressions, and recover an earlier policy when an update degrades behavior. Progress will require evaluation across a sequence of updates, measuring new capability, retention, and safety-relevant failures together.

\paragraph{Broader empirical validation.}
We will extend evaluation to additional benchmarks, robot embodiments, and real-world task distributions. These studies will test transfer beyond RoboCasa365 and to other VLM--action architectures, while providing a setting for adaptive refinement and recovery over longer task horizons. Our goal is a repeatable process for improving embodied behavior from experience.

\section{Conclusion}
We presented \model, a VLA policy that couples training-time representation supervision with recurrent refinement before action generation. IVRR supervises the intermediate features available to the policy, while VAIR refines the conditioning consumed by the DiT; selective joint optimization connects the two. On RoboCasa365, Axiom-0 achieves 62.32\% overall success in our evaluation, exceeding the highest published reference by 4.22 percentage points and Xiaomi-Robotics-1 by 4.92 points, with a 1.7\% increase in mean generation latency under the evaluated setup. This report establishes an initial result for hierarchical self-refinement within 4Axiom Robotics' longer-term research program toward safe recursive self-improvement.

\end{document}